%% file: arxiv.tex
\pdfoutput=1

\documentclass[10pt,twocolumn,letterpaper]{article}

\usepackage[pagenumbers]{cvpr} 

\usepackage{graphicx}
\usepackage{amsmath}
\usepackage{amssymb}
\usepackage{booktabs}
\usepackage{url}
\usepackage{float}
\usepackage{multirow}
\usepackage{bbding}
\usepackage[accsupp]{axessibility}
\usepackage{stfloats}

\usepackage[pagebackref,breaklinks,colorlinks]{hyperref}

\usepackage[capitalize]{cleveref}
\crefname{section}{Sec.}{Secs.}
\Crefname{section}{Section}{Sections}
\Crefname{table}{Table}{Tables}
\crefname{table}{Tab.}{Tabs.}

\def\cvprPaperID{7535} 
\def\confName{CVPR}
\def\confYear{2022}

\begin{document}

\title{SimVP: Simpler yet Better Video Prediction}

\author{Zhangyang Gao$^{1,2,*}$, Cheng Tan$^{1,2,*}$,  Lirong Wu$^{1,2}$, Stan Z. Li$^{1,2}$ \\
$^{1}$ AI Lab, School of Engineering, Westlake University \\
$^{2}$ Institute of Advanced Technology, Westlake Institute for Advanced Study \\
\thanks{Equal contribution}
{\tt\small \{gaozhangyang,tancheng,wulirong,stan.zq.li\}@westlake.edu.cn}
}

\maketitle

\begin{abstract}
  From CNN, RNN, to ViT, we have witnessed remarkable advancements in video prediction, incorporating auxiliary inputs, elaborate neural architectures, and sophisticated training strategies. We admire these progresses but are confused about the necessity: is there a simple method that can perform comparably well? This paper proposes SimVP, a simple video prediction model that is completely built upon CNN and trained by MSE loss in an end-to-end fashion. Without introducing any additional tricks and complicated strategies, we can achieve state-of-the-art performance on five benchmark datasets. Through extended experiments, we demonstrate that SimVP has strong generalization and extensibility on real-world datasets. The significant reduction of training cost makes it easier to scale to complex scenarios. We believe SimVP can serve as a solid baseline to stimulate the further development of video prediction. The code is available at \href{https://github.com/gaozhangyang/SimVP-Simpler-yet-Better-Video-Prediction}{Github}.

\end{abstract}

\begin{table*}[h]
  \centering
  \setlength{\tabcolsep}{5.5mm}{
  \begin{tabular}{ccccc}
    \toprule
            & RNN-RNN-RNN & CNN-RNN-CNN & CNN-ViT-CNN & CNN-CNN-CNN \\
    \midrule
  2014-2015 &  \cite{xingjian2015convolutional, srivastava2015unsupervised,michalski2014modeling}     &  \cite{patraucean2015spatio,lotter2015unsupervised}           &     -     &   \cite{mathieu2015deep}  \\
  2016-2017 &  \cite{lu2017flexible,villegas2017learning,wang2017predrnn,premont2017recurrent}     &   \cite{villegas2017decomposing,finn2016unsupervised,babaeizadeh2017stochastic,liang2017dual,denton2017unsupervised}               &    -    & \cite{van2017transformation,liu2017video,henaff2017prediction} \\
  2018-2019 &  \cite{zhang2019z,sun2019predicting,oliu2018folded,hsieh2018learning,wang2018predrnn++,wang2019memory}   &  \cite{wang2018eidetic,villegas2018hierarchical,denton2018stochastic,castrejon2019improved,yu2019efficient}    &  \cite{weissenborn2019scaling}     &    \cite{gao2019disentangling,kwon2019predicting,xu2018predcnn}  \\
  2020-2021 &  \cite{wang2021predrnn}           &   \cite{hu2020probabilistic,guen2020disentangling}     &  \cite{rakhimov2020latent}         &  \cite{shouno2020photo,chiu2020segmenting}  \\
  \bottomrule 
  \end{tabular}}
  \caption{ Some representative video prediction works since 2014.}
  \label{tab:previous_works}
  \vspace{-2mm}
\end{table*}

\vspace{-3mm}
\section{Introduction}
\label{sec:intro}
A wise person can foresee the future, and so should an intelligent vision model do. Due to spatio-temporal information implying the inner laws of the chaotic world, video prediction has recently attracted lots of attention in climate change \cite{xingjian2015convolutional}, human motion forecasting \cite{babaeizadeh2017stochastic}, traffic flow prediction \cite{wang2019memory} and representation learning \cite{srivastava2015unsupervised}. Struggling with the inherent complexity and randomness of video, lots of interesting works have appeared in the past years. These methods achieve impressive performance gain by introducing novel neural operators like various RNNs \cite{xingjian2015convolutional,wang2019memory,wang2018eidetic,wang2017predrnn,wang2018predrnn++,wang2021predrnn} or transformers \cite{weissenborn2019scaling,rakhimov2020latent}, delicate architectures like autoregressive \cite{ranzato2014video, srivastava2015unsupervised, xingjian2015convolutional, kalchbrenner2017video, ho2019axial} or normalizing flow \cite{yu2019efficient}, and applying distinct training strategies such as adversarial training \cite{mathieu2015deep, saito2017temporal, tulyakov2018mocogan, vondrick2016generating, saito2018tganv2, luc2020transformation, clark2019adversarial, acharya2018towards}. However, there is relatively little understanding of their necessity for good performance since many methods use different metrics and datasets. Moreover, the increasing model complexity further aggravates this dilemma. A question arises: can we develop a simpler model to provide better understanding and performance?

\begin{figure}[h]
  \centering
      \includegraphics[width=0.47\textwidth]{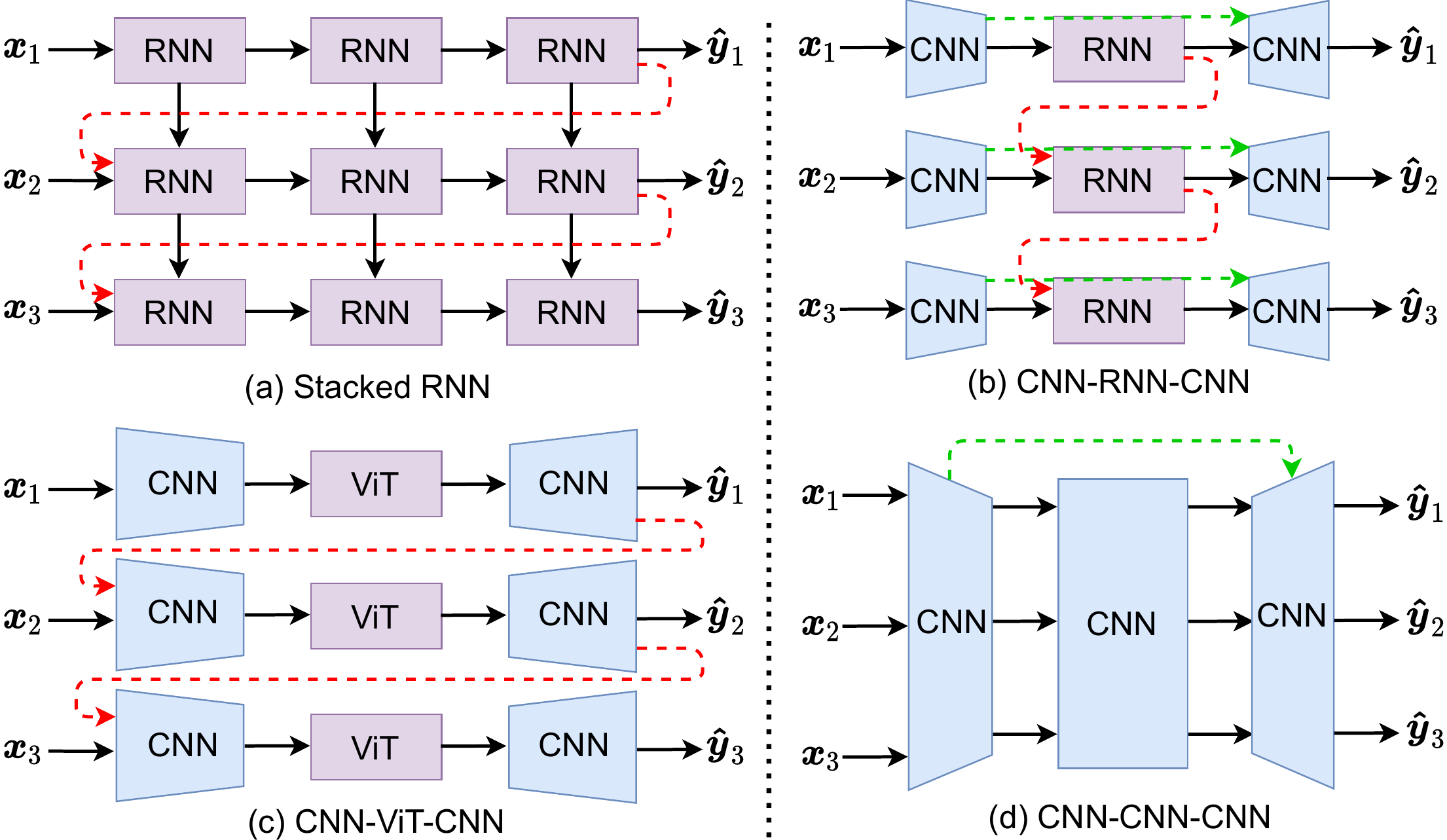}
    \caption{Different architectures for video prediction. Red and blue lines help to learn the temporal evolution and spatial dependency. SimVP belongs to the framework of CNN-CNN-CNN, which can outperform other state-of-the-art methods.}
    \label{fig:cmp_archi}
    \vspace{-2mm}
\end{figure}

Deep video prediction has made incredible progress in the last few years. We divide primary methods into four categories in Figure.~\ref{fig:cmp_archi}, i.e., (1) RNN-RNN-RNN (2) CNN-RNN-CNN (3) CNN-ViT-CNN, and (4) CNN-CNN-CNN. Some representative works are collected in Table.~\ref{tab:previous_works}, from which we observe that RNN models have been favored since 2014. 

In this context, lots of novel RNNs are proposed. ConvLSTM \cite{xingjian2015convolutional} extends fully connected LSTMs to have convolutional structures for capturing spatio-temporal correlations. PredRNN \cite{wang2017predrnn} suggests simultaneously extracting and memorizing spatial and temporal representations. MIM-LSTM \cite{wang2019memory} applies a self-renewed memory module to model both non-stationary and stationary properties. E3D-LSTM \cite{wang2018eidetic} integrates 3D convolutions into RNNs. PhyCell \cite{guen2020disentangling} learns the partial differential equations dynamics in the latent space. 

Recently, vision transformers (ViT) have gained tremendous popularity. AViT \cite{weissenborn2019scaling} merges ViT into the autoregressive framework, where the overall video is divided into volumes, and self-attention is performed within each block independently. Latent AViT \cite{rakhimov2020latent} uses VQ-VAE \cite{oord2017neural} to compress the input images and apply AViT in the latent space to predict future frames. 

In contrast, purely CNN-based models are not as favored as the approaches mentioned above, and fancy techniques are usually required to improve the novelty and performance, e.g., adversarial training \cite{kwon2019predicting}, teacher-student distilling \cite{chiu2020segmenting}, and optical flow \cite{gao2019disentangling}. We admire their significant advancements but expect to exploit how far a simple model can go. In other words, we have made much progress against the baseline results, but have the baseline results been underestimated?

We aim to provide a simpler yet better video prediction model, namely SimVP. This model is fully based on CNN and trained by the MSE loss end-to-end. Without introducing any additional tricks and complex strategies, SimVP can achieve state-of-the-art performance on five benchmark datasets. The simplicity makes it easy to understand and use as a common baseline. The better performance provides a solid foundation for further improvements. We hope this study will shed light on future research.

\begin{figure*}[h]
  \centering
      \includegraphics[width=0.9\textwidth]{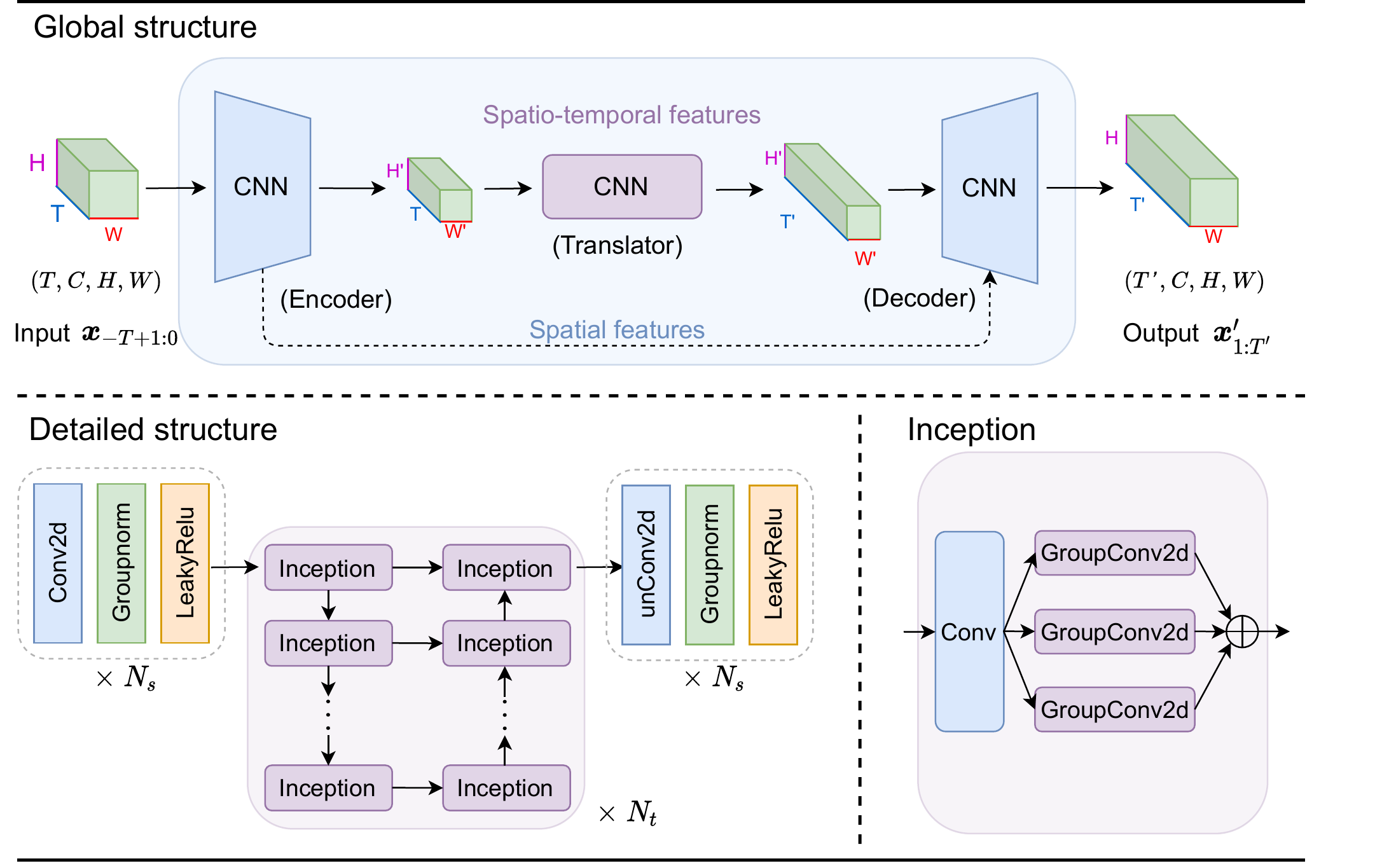}
    \caption{The overall framework of SimVP. Both the Encoder, Translator, and Decoder are built upon CNN. The encoder stacks $N_s$ ConvNormReLU block to extract spatial features, i.e., convoluting $C$ channels on $(H,W)$. The translator employs $N_t$ Inception modules to learn temporal evolution, i.e., convoluting $T \times C$ channels on $(H,W)$. The decoder utilizes $N_s$ unConvNormReLU blocks to reconstruct the ground truth frames, which convolutes $C$ channels on $(H,W)$. } 
    \label{fig:overall_framework}
    \vspace{-4mm}
\end{figure*}


\section{Background}
\paragraph{Problem statement}
Video prediction aims to infer the future frames using the previous ones. Given a video sequence $ \boldsymbol{X}_{t, T}=\{ \boldsymbol{x}_i \}_{t - T + 1}^{t}$ at time $t$ with the past $T$ frames, our goal is to predict the future sequence $\boldsymbol{Y}_{t, T'} = \{ \boldsymbol{x}_i \}_{t}^{t + T'}$ at time $t$ that contains the next $T'$ frames, where $\boldsymbol{x}_i \in \mathbb{R}^{C, H, W}$ is an image with channels $C$, height $H$, and width $W$. Formally, the predicting model is a mapping $\mathcal{F}_{\Theta}: \boldsymbol{X}_{t, T} \mapsto \boldsymbol{Y}_{t, T'}$ with learnable parameters $\mathbf{\Theta}$, optimized by:
\begin{equation}
    \Theta^* = \arg\min_{\Theta} \mathcal{L}(\mathcal{F}_{\Theta}(\boldsymbol{X}_{t, T}), \boldsymbol{Y}_{t, T'})
\end{equation}
where $\mathcal{L}$ can be various loss functions, and we simply employ MSE loss in our setting.

\paragraph{RNN-RNN-RNN} As shown in Figure.~\ref{fig:cmp_archi} (a), this kind of method stacks RNN to make predictions. They usually design novel RNN modules (local) and overall architectures (global). Recurrent Grammar Cells \cite{michalski2014modeling} stacks multiple gated autoencoders in a recurrent pyramid structure. ConvLSTM \cite{xingjian2015convolutional} extends fully connected LSTMs to have convolutional computing structures to capture spatio-temporal correlations. PredRNN \cite{wang2017predrnn} suggests simultaneously extracting and memorizing spatial and temporal representations. PredRNN++ \cite{wang2018predrnn++} proposes gradient highway unit to alleviate the gradient propagation difficulties for capturing long-term dependency. MIM-LSTM \cite{wang2019memory} uses a self-renewed memory module to model both the non-stationary and stationary properties of the video. dGRU \cite{oliu2018folded} shares state cells between encoder and decoder to reduce the computational and memory costs. Due to the excellent flexibility and accuracy, these methods play fundamental roles in video prediction.

\paragraph{CNN-RNN-CNN} This framework projects video frames to the latent space and employs RNN to predict the future latent states, seeing Figure.~\ref{fig:cmp_archi} (b). In general, they focus on modifying the LSTM and encoding-decoding modules. Spatio-Temporal video autoencoder \cite{patraucean2015spatio} incorporates ConvLSTM and an optical flow predictor to capture changes over time. Conditional VRNN \cite{castrejon2019improved} combines CNN encoder and RNN decoder in a variational generating framework. E3D-LSTM \cite{wang2018eidetic} applies 3D convolution for encoding and decoding and integrates it in latent RNNs for obtaining motion-aware and short-term features. CrevNet \cite{yu2019efficient} proposes using CNN-based normalizing flow modules to encode and decode inputs for information-preserving feature transformations. PhyDNet \cite{guen2020disentangling} models physical dynamics with CNN-based PhyCells. Recently, this framework has attracted considerable attention, because the CNN encoder can extract decent and compressed features for accurate and efficient prediction.

\paragraph{CNN-ViT-CNN} This framework introduces Vision Transformer (ViT) to model latent video dynamics. By extending language transformer \cite{vaswani2017attention} to ViT \cite{dosovitskiy2020image}, a wave of research has been sparked recently. As to image transformer, DeiT \cite{touvron2021training} and Swin Transformer \cite{liu2021swin} have achieved state-of-the-art performance on various vision tasks. The great success of image transformer has inspired the investigation of video transformer. VTN \cite{neimark2021video} applies sliding window attention on temporal dimension following a 2D spatial feature extractor. TimeSformer and ViViT \cite{bertasius2021space, arnab2021vivit} study different space-time attention strategies and suggest that separately applying temporal and spatial attention can achieve superb performance. MViT \cite{fan2021multiscale} extracts multiscale pyramid features to provide state-of-the-art results on SSv2. Video Swin Transformer \cite{liu2021video} expands Swin Transformer from 2D to 3D, where the shiftable local attention schema leads to a better speed-accuracy trade-off. Most of the models above are designed for video classification; works about video prediction \cite{weissenborn2019scaling,rakhimov2020latent} using ViT are still limited. More related works may emerge in the future.

\vspace{-3mm}
\paragraph{CNN-CNN-CNN} The CNN-based framework is not as popular as the previous three because it is so simple that complex modules and training strategies are usually required. DVF \cite{liu2017video} suggests learning the voxel flow by CNN autoencoder to reconstruct a frame by borrowing voxels from nearby frames. PredCNN \cite{xu2018predcnn} combines cascade multiplicative units (CMU) with CNN to capture inter-frame dependencies. DPG \cite{gao2019disentangling} disentangles motion and background via a flow predictor and a context generator. \cite{chiu2020segmenting} encodes RGB frames from the past and decodes the future semantic segmentation by using CNN and teacher-student distilling. \cite{shouno2020photo} uses a hierarchical neural model to make predictions at different spatial resolutions and train the model with adversarial and perceptual loss functions. While these approaches have made progress, we are curious about what happens if the complexity is reduced. Is there a solution that is are much simpler but can exceed or match the performance of state-of-the-art methods?

\vspace{-3mm}
\paragraph{Motivation} We have witnessed many terrific methods that have achieved outstanding performance. However, as the models become more complex, understanding their performance gain is an inevitable challenge, and scaling them into large datasets is intractable. This work does not propose new modules. Instead, we aim to build a simple network based on existing CNNs and see how far the simple model can go in video prediction. 

\section{SimVP}
\label{sec:method}
Our model, dubbed \textit{SimVP}, consists of an encoder, a translator and a decoder built on CNN, seeing Figure.~\ref{fig:overall_framework}. The encoder is used to extract spatial features, the translator learns temporal evolution, and the decoder integrates spatio-temporal information to predict future frames. 

\vspace{-3mm}
\paragraph{Encoder} The encoder stacks $N_s$ ConvNormReLU blocks (Conv2d+LayerNorm+LeakyReLU) to extract spatial features, i.e., convoluting $C$ channels on $(H,W)$. The hidden feature is:

\vspace{-5mm}
\begin{equation}
  \centering
      \label{eq:encoder_layernorm}
      z_{i} = \sigma(\mathrm{LayerNorm} (\mathrm{Conv2d}(z_{i-1}))),  1 \leq i \leq N_s 
\end{equation}
where the input $z_{i-1}$ and output $z_{i}$ shapes are $(T,C,H,W)$ and $(T,\hat{C},\hat{H},\hat{W})$, respectively.

\paragraph{Translator} The translator employs $N_t$ Inception modules to learn temporal evolution, i.e., convoluting $T \times C$ channels on $(H,W)$. The Inception module consists of a bottleneck Conv2d with $1 \times 1$ kernel followed by parallel GroupConv2d operators. The hidden feature is:

\begin{equation}
  \centering
      \label{eq:encoder_inception}
      z_{j} = \mathrm{Inception}( z_{j-1} ),  N_s < j \leq N_s+N_t
\end{equation}
where the inputs $z_{j-1}$ and output $z_{j}$ shapes are $(T \times C,H,W)$ and $(\hat{T} \times \hat{C},H,W)$.

\paragraph{Decoder} The decoder utilizes $N_s$ unConvNormReLU blocks (ConvTranspose2d+GroupNorm+LeakyReLU) to reconstruct the ground truth frames, which convolutes $C$ channels on $(H,W)$. The hidden feature is:

\begin{equation}
  \centering
  \begin{aligned}
  \label{eq:decoder_layernorm}
    z_{k} = \sigma(\mathrm{GroupNorm} (\mathrm{unConv2d}(z_{k-1}))),\\ N_s+N_t<k \leq 2N_s+N_t
  \end{aligned}
\end{equation}
where the shapes of input $z_{k-1}$ and output $z_{k}$ are $(T,\hat{C},\hat{H},\hat{W})$ and $(T,C,H,W)$, respectively. We use ConvTranspose2d \cite{dumoulin2016guide} to serve as the $\mathrm{unConv2d}$ operator.

\paragraph{Summary} SimVP does not use advanced modules such as RNN, LSTM and Transformer, nor introduce complex training strategies such as adversarial training and curriculumn learning. All the things we need are CNN, shortcuts and vanilla MSE loss.

\section{Experiments}

\paragraph{Metrics} We employ MSE, MAE, Structural Similarity Index Measure (SSIM), and Peak Signal to Noise Ratio (PSNR) to evaluate the quality of predictions, following \cite{yu2019efficient,guen2020disentangling,oprea2020review}. We also report the running time per epoch and the memory footprint per sample on a single NVIDIA-V100 to provide a comprehensive view for future research. \footnote{The realistic running time is more reliable to FLOPs or MACs of the model, e.g., when RNN and CNN have the same FLOPs, RNN takes much more time due to its recurrent computation. }

\paragraph{Datasets} We conduct experiments on five datasets for evaluation. The statistics are summarized in Table.~\ref{Table:dataset}, including the number of training samples $n_{train}$, number of testing samples $n_{test}$, image resolution $(C,H,W)$, input sequence length $T$ and forecasting sequence length $T'$. The detailed dataset description can be found in the appendix.

\begin{table}[h]
  \centering
  \caption{The statistics of datasets. The training or testing set has $N_{train}$ or $N_{test}$ samples, each of which consists $T$ or $T'$ images with the shape $(C,H, W)$.}
  \resizebox{\columnwidth}{!}{
      \begin{tabular}{cccccc}
      \toprule
                 & $N_{train}$ & $N_{test}$ & $(C,H, W)$ & $T$ & $T'$ \\
      \midrule
      MMNIST     &  10000        &  10000       & (1, 64, 64)   & 10  & 10 \\
      TrafficBJ &  19627        &  1334       & (2, 32, 32)   & 4   & 4 \\
      Human3.6  &  2624         &   1135      & (3, 128, 128) & 4   & 4 \\
      Caltech Pedestrian & 2042   &1983       & (3, 128, 160)   & 10    & 1\\
      KTH        &  5200         & 3167         & (1, 128, 128) & 10  & 20 or 40\\
      \bottomrule
      \end{tabular}}
  
      \label{Table:dataset}
\end{table}

\vspace{-3mm}
\subsection{How far can SimVP go?}
We pursue the simple but effective model. The simplicity has been described in Section.~\ref{sec:method}. The effectiveness will be verified through exploratory experiments in this section.

\vspace{-3mm}
\paragraph{Challenge} We aim to provide a comprehensive and rigorous view of video prediction methods. However, three challenges stand in our way:

\begin{itemize}
  \item Various methods may adopt disparate metrics.
  \setlength{\itemsep}{1pt}
  \setlength{\parsep}{1pt}
  \setlength{\parskip}{1pt}
  \item These methods apply experiments on different datasets with distinct protocols.
  \setlength{\itemsep}{1pt}
  \setlength{\parsep}{1pt}
  \setlength{\parskip}{1pt}
  \item They use different code frameworks and unique tricks, making it difficult to compare fairly.
\end{itemize}

\vspace{-5mm}
\paragraph{Solution} To overcome aforemesioned challenges, we choose the common used dataset (Moving MNIST) and metrics (MSE and SSIM) to evaluate recent important researches. We directly report the best metrics according to the original papers, avoiding the risk of performance degradation caused by our reproduction. Any method that does not use the same dataset, metrics, or protocol will be neglected here. For convenience, we provide the publication status and Github links of these method in Table.~\ref{tab:cmp_performance_mmnist}.

\vspace{-2mm}
\paragraph{Results and Discovery} As shown in Table.~\ref{tab:cmp_performance_mmnist}, SimVP achieves state-of-the-art MSE and SSIM on Moving MNIST. We observe that SimVP, PhyDNet, and CrevNet significantly outperform previous methods, with MSE reduction up to 42\%. However, SimVP is much simpler than PhyDNet and CrevNet, without using RNN, LSTM, or complicated modules, which are considered as the important reason for performance improvement. Through these explorations, we are excited to find that \textbf{it is promising to achieve better performance with a extremely simple model.} Perhaps previous works pay too much attention to the model complexity and novelty, and it's time to go back to basics because a simpler model makes things clearer.



\begin{table}[h]
  \small
  \centering
  \caption{Performance comparision of various methods on Moving MNIST. The source code of SimVP will be released soon.}

  \resizebox{\columnwidth}{!}{
  \begin{tabular}{ccccc}
  \toprule
  Method   & Conference  & MSE     & SSIM  & Github\\
  \midrule
  ConvLSTM \cite{xingjian2015convolutional} & (NIPS 2015)   & 103.3  & 0.707  & \href{https://github.com/ndrplz/ConvLSTM_pytorch}{PyTorch}\\
  PredRNN \cite{wang2017predrnn} & (NIPS 2017)   & 56.8   & 0.867  & \href{https://github.com/thuml/predrnn-pytorch}{PyTorch}\\
  PredRNN-V2 \cite{wang2021predrnn}  & (Arxiv 2021) & 48.4     & 0.891 & \href{https://github.com/thuml/predrnn-pytorch}{PyTorch}\\
  CausalLSTM \cite{wang2018predrnn++} & (ICML 2018) & 46.5   & 0.898 & \href{https://github.com/Yunbo426/predrnn-pp}{Tensorflow}\\
  MIM \cite{wang2019memory} & (CVPR 2019)       & 44.2   & 0.910 & \href{https://github.com/Yunbo426/MIM}{Tensorflow}\\
  E3D-LSTM \cite{wang2018eidetic} & (ICLR 2018)  & 41.3   & 0.920 & \href{https://github.com/google/e3d_lstm}{Tensorflow}\\
  PhyDNet \cite{guen2020disentangling} & (CVPR 2020)   & 24.4    & 0.947 & \href{https://github.com/vincent-leguen/PhyDNet}{PyTorch}\\
  CrevNet \cite{yu2019efficient}  & (ICLR 2020) & 22.3  & 0.949 & \href{https://github.com/rrxi/CrevNet}{PyTorch}\\
  \midrule
  SimVP  & --    & 23.8   & 0.948 & PyTorch\\
  \bottomrule
  \end{tabular}}
  \label{tab:cmp_performance_mmnist}
\end{table}

\vspace{-5mm}
\paragraph{Simplicity leads to efficiency.} Another benefit that comes from simplicity is good computational efficiency. In Table.~\ref{tab:cmp_computation_mmnist}, we compare GPU memory (per sample), FLOPs (per image) and training time of SOTA methods on Moving MNIST. As CNN has good computational optimization and avoids iterative calculation, the training process of SimVP is much faster than others, which means that SimVP can be used and extended more easily. 

\begin{table}[h]
  \small
  \centering
  \caption{Computation comparision on Moving MNIST. We report the per sample memory overhead, per frame FLOPs, and total training time. For methods marked with $*$, we report the results reproduced by their official codes. Other results refer to \cite{yu2019efficient}. }
  
  \resizebox{\columnwidth}{!}{
  \begin{tabular}{cccc}
  \toprule
  Method     & Memory & FLOPs  & Training time\\
  \midrule
  ConvLSTM   & 1043MB  & 107.4G  & --\\
  PredRNN    & 1666 MB & 192.9 G & --\\
  CausalLSTM & 2017 MB & 106.8 G & --\\
  E3D-LSTM   & 2695 MB & 381.3 G & --\\
  CrevNet $^*$    & 224 MB  & 1.652 G & $\approx 10d$ (300k iters)\\
  PhyDNet $^*$   & 200 MB  & 1.633 G & $\approx 10d$ (2k epochs)\\
  \midrule
  SimVP $^*$     & 412 MB  & 1.676 G & $\approx 2d$ (2k epochs)\\ 
  \bottomrule
  \end{tabular}}
  \label{tab:cmp_computation_mmnist}
\end{table}

\vspace{-2mm}
\subsection{ Translator: should we use RNN, Transformer or CNN?}
With the rapid development of novel temporal modules based on RNN, Transformer, and CNN, researchers may be dazzled and confused about which one to choose for video modeling. In SimVP, the Translator module is responsible for learning temporal evolution. We replace the CNN-based Translator with the most representative RNNs and Transformers to reveal that which time module is more suitable for video modeling under the SimVP framework.

\vspace{-3mm}
\paragraph{Translator selection} The CNN-based Translator serves as the baseline, since it is quite simple yet effective. We hope to find new modules that is promising to outperform the Translator to inspire subsequent researches. For RNN, we choose currently state-of-the-art \textbf{PhyDNet} \cite{guen2020disentangling} and \textbf{CrevNet} \cite{yu2019efficient}. 
As suggested in PhyDNet, we use the PhyCell, a novel time module considering physical dynamics, for temporal modeling. As to CrevNet, we use normalizing flow autoencoder + ST-LSTM \cite{wang2017predrnn} as the translator. For Transformers, we chose recent influential work such as Video Swin Transformer \cite{liu2021video} and Latent Video Transformer \cite{weissenborn2019scaling,rakhimov2020latent}. To make these modules workable under the SimVP framework, we may modify a few implementation details without changing the core algorithm, such as removing the autoregressive generation scheme and re-implementing position encoding.

\vspace{-2mm}
\paragraph{Setting} We compare five translators on Moving MNIST and Human, keeping the encoder and decoder the same. We adjust the hyperparameters of these translators to make them work with similar GPU memory footprints. By default, we use batch size 16, epoch 100, and Adam optimizer. The number of encoder and decoder layers is 4. We choose the largest learning rate from $\{1e^{-2},1e^{-3},1e^{-4}\}$ under the premise of stable training. 



\begin{table*}[h]
  \small
  \centering
      \caption{ SimVP vs SOTA. The optimal(or suboptimal) results are marked by \textbf{bold}(or \underline{underlined}).}
      \resizebox{2 \columnwidth}{!}{
          \begin{tabular}{c ccc ccc ccc}
          \toprule
                    & \multicolumn{3}{c}{Moving MNIST} & \multicolumn{3}{c}{TrafficBJ} & \multicolumn{3}{c}{Human3.6} \\
                    & MSE$\downarrow$    & MAE$\downarrow$   & SSIM$\uparrow$     & MSE $\times$ 100 $\downarrow$ & MAE$\downarrow$   & SSIM$\uparrow$    & MSE / 10 $\downarrow$ & MAE / 100 $\downarrow$ & SSIM$\uparrow$   \\ 
          \midrule
          ConvLSTM   & 103.3  & 182.9 &  0.707             & 48.5  & 17.7  & 0.978             & 50.4 & 18.9 & 0.776             \\
          PredRNN    & 56.8   & 126.1 &  0.867              & 46.4  & 17.1  & 0.971             & 48.4 & 18.9 & 0.781             \\
          CausalLSTM & 46.5   & 106.8 &  0.898              & 44.8  & 16.9  & 0.977             & 45.8 & 17.2 & 0.851             \\
          MIM        & 44.2   & 101.1 &  0.910              & 42.9  & 16.6  & 0.971             & 42.9 & 17.8 & 0.790             \\
          E3D-LSTM   & 41.3   & 86.4  &  0.910              & 43.2  & 16.9  & 0.979             & 46.4 & 16.6 & 0.869             \\
          PhyDNet    & \underline{24.4}   & \underline{70.3}  &  \underline{0.947}              & \underline{41.9}  & \textbf{16.2}  & \textbf{0.982}             & \underline{36.9} & \underline{16.2} & \underline{0.901}             \\ \hline
          SimVP      & \textbf{23.8}   & \textbf{68.9}  &  \textbf{0.948}    & \textbf{41.4}  & \textbf{16.2}  & \textbf{0.982}          & \textbf{31.6} & \textbf{15.1} &  \textbf{0.904}            \\ 
          \bottomrule
          \end{tabular}
      }
  \label{table:basic_sota}
\end{table*}

\begin{figure}[h]
  \centering
    \includegraphics[width=0.47\textwidth]{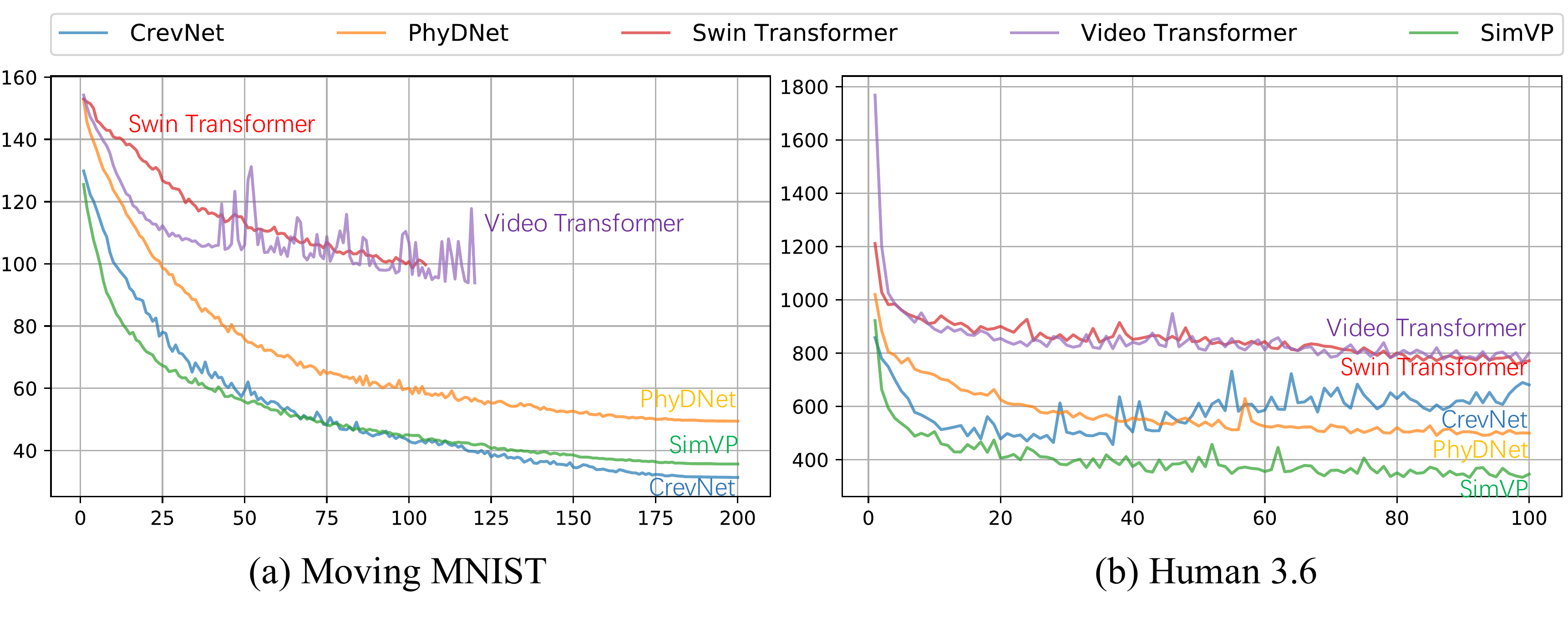}
    \caption{Training dynamics of translators on Moving MNIST and Human3.6. The x-axis is the training epochs and the y-axis is the evaluated MSE. SimVP converges faster than other methods in the early training phase, but CrevNet performs better in the long run.}
    \label{fig:cmp_translators_merge}
\end{figure}

\vspace{-5mm}
\paragraph{Results and Discoveries} In Figure.~\ref{fig:cmp_translators_merge}, we show the training dynamics using various translators on Moving MNIST and Human3.6, respectively.  CrevNet looks like a good solution because the inherent simplicity of Moving MNIST, while our SimVP is suboptimal. However, this does not always hold true when it comes to real-world datasets. For example, on Human dataset (Figure.~\ref{fig:cmp_translators_merge}(b)), SimVP significantly outperforms other methods. As to the training stability, CNN performs better than RNN (Figure.~\ref{fig:cmp_translators_lr}, appendix), since it will not fluctuate violently under large learning rate. In short, our discoveries are:

\begin{enumerate}
  \item CNN and RNN achieves state-of-the-art performance under limited computation costs. 
  \setlength{\itemsep}{1pt}
  \setlength{\parsep}{1pt}
  \setlength{\parskip}{1pt}
  \item RNN converges faster than others in the long run (Figure.~\ref{fig:cmp_translators_merge}) if the model capacity is sufficient.

  \setlength{\itemsep}{1pt}
  \setlength{\parsep}{1pt}
  \setlength{\parskip}{1pt}
  \item CNN training is more robust and does not fluctuate dramatically at large learning rates (Figure.~\ref{fig:cmp_translators_lr}).

  \setlength{\itemsep}{1pt}
  \setlength{\parsep}{1pt}
  \setlength{\parskip}{1pt}
  \item Transformer has no advantage in our SimVP framework under the similar resource consumption.
\end{enumerate}


\subsection{Does SimVP achieve SOTA on general cases?}
From previous analysis, we believe that SimVP has potential to outperform recent state-of-the-art methods. In this section, we provide more experimental evidence to confirm this claim. Specifically, we aim to answer three questions:

\begin{itemize}
  \item \textbf{Q1:} Can SimVP achieve SOTA results on other common benchmarks?
  \setlength{\itemsep}{1pt}
  \setlength{\parsep}{1pt}
  \setlength{\parskip}{1pt}
  
  \item \textbf{Q2:} Does SimVP generalize well across different datasets?
  \setlength{\itemsep}{1pt}
  \setlength{\parsep}{1pt}
  \setlength{\parskip}{1pt}
  \item \textbf{Q3:} Does SimVP extend well to the case of flexible predictive length?
\end{itemize}

\vspace{-5mm}
\paragraph{Setting of Q1} Reporting the widely used metrics on standard benchmark datasets is the key to advancing the research progress. We evaluate SimVP on three common used benchmarks, i.e., Moving MNIST, TrafficBJ and Human3.6.  Moving MNIST \cite{srivastava2015unsupervised} consists of two digits independently moving within the 64 $\times$ 64 grid and bounced off the boundary. By assigning different initial locations and velocities to each digit, we can get an infinite number of sequences, and predict the future 10 frames from previous 10 frames. TrafficBJ contains the trajectory data in Beijing collected from taxicab GPS with two channels, i.e. inflow or outflow defined in \cite{zhang2017deep}.  Following \cite{wang2019memory}, we transform the data into $[0, 1]$ via max-min normalization. Since the original data is between -1 and 1, the reported MSE and MAE are 1/4 and 1/2 of the original ones, consistent with previous literature \cite{guen2020disentangling,wang2019memory}. Models are trained to predict 4 next frames by observing prioring 4 frames. Human3.6 \cite{ionescu2013human3} is a complex human pose dataset with 3.6 million samples, recording different activities. Similar to \cite{srivastava2015unsupervised,wang2019memory,guen2020disentangling}, only videos with "walking" scenario are used, and 4 future frames are generated by feeding previous 4 RGB frames. Following \cite{guen2020disentangling}, we report the MSE, MAE and SSIM of SimVP in Table. \ref{table:basic_sota}. Six state-of-the-art RNN baselines are chosen for comparison, including ConvLSTM \cite{xingjian2015convolutional}, PredRNN \cite{wang2017predrnn}, CausalLSTM \cite{wang2018predrnn++}, MIM \cite{wang2019memory}, E3D-LSTM \cite{wang2018eidetic} and PhyDNet \cite{guen2020disentangling}. We train SimVP on Moving MNIST, Human3.6, and TrafficBJ for 2k, 100 and 80 epochs, respectively. We use Adam optimizer and the learning rate is 0.01.

\begin{figure}[h]
  \centering
      \includegraphics[width=3in]{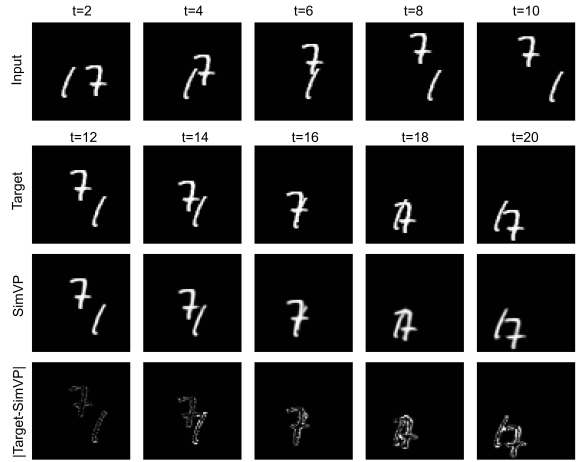}
    \caption{ Visualization of Moving MNIST.}
    \label{fig:mmnist_8317}
\end{figure}

\vspace{-5mm}
\paragraph{Answer of Q1} SimVP can achieve SOTA results on lightweight benchmarks. From Table. \ref{table:basic_sota}, we observe that SimVP outperform all RNN baselines in all settings. The improvements on Moving MNIST and TrafficBJ are modest;  On Human3.6, the relative improvement is significant. In addition, SimVP takes training time compared to previous PhyDNet. Both performance and computing advantages support our answer.


\vspace{-2mm}
\paragraph{Setting of Q2} Generalizing the knowledge across different datasets, especially under the unsupervised setting, is the core research point of machine learning. To investigate the generalization ability of SimVP, we train the model for 50 epochs on KITTI and evaluate it on Caltech Pedestrian. KITTI \cite{geiger2013vision} is one of the most popular datasets for mobile robotics and autonomous driving. It includes hours of traffic scenarios recorded with high-resolution RGB images. CalTech Pedestrian \cite{dollarCVPR09peds} is a driving dataset focused on detecting pedestrians. It is conformed of approximately 10 hours of $640 \times 480$ 30 FPS video taken from a vehicle driving through regular traffic in an urban environment. Models are trained on KITTI dataset to predict the next frame after 10-frame warm-up and are evaluated on Caltech Pedestrian. Compared with the previous experiments, carmounted camera videos dataset and the distinct training-evaluating data present another level of difficulty for video prediction as it describes various nonlinear three-dimensional dynamics of multiple moving objects including backgrounds. Following \cite{lotter2016deep,yu2019efficient,oprea2020review}, we center-crop  Sixteen baselines are selected for comparision, including BeyondMSE \cite{mathieu2015deep}, MCnet\cite{villegas2017decomposing}, DVF\cite{liu2017video}, Dual-GAN\cite{liang2017dual}, CtrlGen\cite{hao2018controllable},PredNet \cite{lotter2016deep}, ContextVP\cite{byeon2018contextvp}, GAN-VGG, G-VGG, G-MAE, GAN-MAE\cite{shouno2020photo}, SDC-Net \cite{reda2018sdc}, rCycleGan \cite{kwon2019predicting},  DPG \cite{gao2019disentangling}, CrevNet \cite{yu2019efficient} and STMFANet \cite{jin2020exploring}. 

\begin{table}[h]
  \small
  \centering
  \caption{Results on Caltech Pedestrian dataset.}
  \setlength{\tabcolsep}{3.5mm}{
     \begin{tabular}{cccc}
     \toprule
                  & \multicolumn{3}{c}{Caltech Pedestrian ($10 \rightarrow 1$)} \\ \cline{2-4} 
     Method       & MSE$\downarrow$    & SSIM$\uparrow$    & PSNR$\uparrow$  \\
     \midrule
     BeyondMSE \cite{mathieu2015deep}   & 3.42   & 0.847   & -      \\
     MCnet   \cite{villegas2017decomposing}     & 2.50   & 0.879   & -      \\
     DVF    \cite{liu2017video}      & -      & 0.897   & 26.2   \\
     Dual-GAN  \cite{liang2017dual}   & 2.41   & 0.899   & -       \\
     CtrlGen  \cite{hao2018controllable}    & -      & 0.900   & 26.5   \\
     PredNet   \cite{lotter2016deep}   & 2.42   & 0.905   & 27.6   \\
     ContextVP \cite{byeon2018contextvp}   & 1.94   & 0.921   & 28.7   \\
     GAN-VGG  \cite{shouno2020photo}    & -      & 0.916   & -      \\
     G-VGG   \cite{shouno2020photo}     & -      & 0.917   & -     \\
     SDC-Net   \cite{reda2018sdc}   & 1.62   & 0.918   & -      \\
     rCycleGan  \cite{kwon2019predicting}  & \underline{1.61}   & 0.919   & 29.2  \\
     DPG    \cite{gao2019disentangling}      & -      & 0.923   & 28.2   \\
     G-MAE  \cite{shouno2020photo}      & -      & 0.923   & -      \\
     GAN-MAE  \cite{shouno2020photo}    & -      & 0.923   & -      \\
     CrevNet  \cite{yu2019efficient}    & -      & 0.925   & \underline{29.3}  \\
     STMFANet \cite{jin2020exploring}    & -      & \underline{0.927}   & 29.1   \\ 
     \midrule
     SimVP (ours) & \textbf{1.56}   & \textbf{0.940}   & \textbf{33.1} \\
     \bottomrule
     \label{table:Caltech}
     \end{tabular}}
\end{table}

\vspace{-8mm}
\begin{figure}[h]
  \centering
      \includegraphics[width=3in]{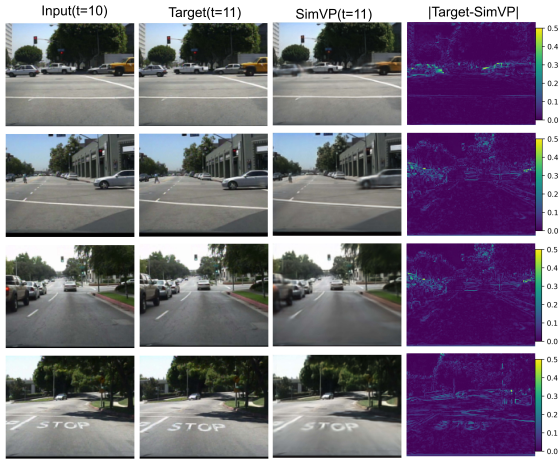}
    \caption{ Visualization of Caltech $(10 \rightarrow 1)$.}
    \label{fig:Caltech}
\end{figure}

\vspace{-3mm}
\paragraph{Answer of Q2} SimVP generalize well cross different datasets, seeing Figure.~\ref{fig:Caltech}. Although there is room to improve the clarity of the generated objects, the evaluated metrics have exceeded previous methods. As shown in Table. \ref{table:Caltech}, the MSE is reduced by 3.1\%, the SSIM is improved by 1.4\%, and the PSNR is improved by 13.0\%. The training process can be finished within 4h.



\begin{table*}[t]
  \footnotesize
  \centering
  \setlength{\tabcolsep}{1.8mm}{
  \begin{tabular}{c|ccccc|ccccc|c}
  \hline
                       &    & model 1   & model 2   & model 3   & model 4   & model 5   & model 6   & model 7   & model 8  & model 9 & SimVP   \\ \hline
  \multirow{4}{*}{ \rotatebox{90}{Archi}} & S-UNet    & --   & \Checkmark   & \Checkmark   & \Checkmark    & \Checkmark    & \Checkmark   & \Checkmark   & \Checkmark    & \Checkmark    & \Checkmark  \\
                                      & T-UNet     & \Checkmark   & --   & \Checkmark   & \Checkmark   & \Checkmark   & \Checkmark   & \Checkmark  & \Checkmark    & \Checkmark   & \Checkmark   \\
              & \#Groups         & 4  & 4    & 1    & 4    & 4    & 4     & 4     & 4    & 4   & 4      \\
                                      & Norm           & G & G & G & B & G & G & G & G & G & G \\ \hline
  \multirow{6}{*}{\rotatebox{90}{Kernel}}    & (3) +$c_{t}$        & --   & --    & --     & --    & \Checkmark    & --     & --     & --    & --    & --        \\
                                      & (5)+$c_{t}$         & --   & --   & --   & --   & --   & \Checkmark    & --     & --  & --   & --   \\
                                      & (7)+$c_{t}$            & --   & --   & --   & --   & --   & --   & \Checkmark   & --   & --  & --  \\
                                      & (11)+$c_{t}$           & --        & --        & --    & --  & --  & --  & --        & \Checkmark    & --   & --        \\
                                      & (11)+$2c_{t}$ & --        & --        & --  & --  & --   & --    & --  & --    & \Checkmark    & --    \\
                                      & (3,5,7,11)+$c_{t}$     & \Checkmark       & \Checkmark    & \Checkmark  & \Checkmark & --     & --   & --        & --    & --    & \Checkmark       \\ \hline
  \multirow{3}{*}{\rotatebox{90}{MSE}}& Moving MNIST         & 41.7  & 41.5  & 44.8  & 41.0   & 58.9   &  51.1 & 49.1 &  46.3   & 44.8   &  41.7     \\
                                      & TrafficBJ $(\times 100)$     & 43.5     & 41.5   &  44.5   &  44.3   &  43.5   &  44.0   &  43.3  &  42.3   &  42.3  &  42.0  \\
                                      & Human3.6 ($/ 10$)         & 32.4  & 33.4  & 33.0  & 34.8 & 37.3 & 34.0  & 32.9  &  33.4  &  32.1  & 32.0\\ \hline
      & Summary & $\downarrow $  & $\downarrow $  & $\downarrow \downarrow \downarrow$ & $\downarrow$ & $\downarrow \downarrow \downarrow$ & $\downarrow \downarrow \downarrow$ & $\downarrow \downarrow \downarrow$ & $\downarrow \downarrow$ & $\downarrow$ & --\\\hline
  \end{tabular}}
  \caption{Ablation study. S-UNet or T-UNet denotes the shortcut connection in the spatial or temporal encoder-decoder. \#Groups is the number of convolutional groups. G and B indicate group normalization and batch normalization. (3,5,7,11)+$c_{t}$ means the Conv kernels of the Inception module plus the translator's hidden dimension. Note that the spatial Enc and Dec are fixed on each dataset, seeing Table.~\ref{tab:hyper_param} (appendix) for detailed settings. Finally, we compare models with SimVP, and results with a gap of less than 0.5 are regarded as the same.}
  \label{tab:ablation}
  \vspace{-3mm}
\end{table*}

\vspace{-2mm}
\paragraph{Setting of Q3} A possible limitation of CNN-based methods is that it may be diffifcult to scale to prediction with flexible length. We handle this problem by imitating RNN, that is taking previous predictions as recent inputs to recursively produce long-term predictions. Following \cite{wang2018eidetic,oprea2020review}, we compare the PSNR and SSIM of SimVP with other baselines on KTH, seeing Table.\ref{table:KTH}, where we train SimVP for 100 epochs. The KTH dataset \cite{schuldt2004recognizing} contains 25 individuals performing 6 types of actions, i.e., walking, jogging, running, boxing, hand waving and hand clapping. Following \cite{villegas2017decomposing,wang2018eidetic}, we use person 1-16 for training and 17-25 for testing. Models are trained to predict next 20 or 40 frames from the previous 10 observations. Sixteen baselines are included, such as MCnet\cite{villegas2017decomposing}, ConvLSTM \cite{xingjian2015convolutional}, SAVP, SAVP-VAE\cite{lee2018stochastic}, VPN \cite{kalchbrenner2017video}, DFN \cite{jia2016dynamic}, fRNN \cite{oliu2018folded}, Znet \cite{zhang2019z}, SV2P \cite{babaeizadeh2017stochastic}, PredRNN \cite{wang2017predrnn}, VarNet \cite{jin2018varnet}, PredRNN++ \cite{wang2018predrnn++}, MSNET \cite{lee2018mutual}, E3d-LSTM \cite{wang2018eidetic} and STMFANet \cite{jin2020exploring}. 

\begin{table}[H]
  \small
  \centering
      \caption{Results on KTH dataset. }
      \setlength{\tabcolsep}{1.5mm}{
          \begin{tabular}{ccccc}
          \toprule
                        & \multicolumn{2}{c}{KTH ($10 \rightarrow 20$)} & \multicolumn{2}{c}{KTH ($10 \rightarrow 40$)}\\ \cline{2-5} 
          Method       & SSIM$\uparrow$         & PSNR$\uparrow$        & SSIM$\uparrow$        & PSNR$\uparrow$    \\
          \midrule
          MCnet \cite{villegas2017decomposing}       & 0.804      & 25.95     & 0.73   & 23.89\\
          ConvLSTM \cite{xingjian2015convolutional}    & 0.712        & 23.58       & 0.639       & 22.85      \\
          SAVP  \cite{lee2018stochastic}       & 0.746        & 25.38       & 0.701       & 23.97       \\
          VPN  \cite{kalchbrenner2017video}        & 0.746        & 23.76       & --          & --        \\
          DFN  \cite{jia2016dynamic}        & 0.794        & 27.26       & 0.652       & 23.01     \\
          fRNN  \cite{oliu2018folded}       & 0.771        & 26.12       & 0.678       & 23.77     \\
          Znet  \cite{zhang2019z}       & 0.817        & 27.58       & --           & --       \\
          SV2Pi \cite{babaeizadeh2017stochastic}     & 0.826        & 27.56       & 0.778       & 25.92     \\
          SV2Pv  \cite{babaeizadeh2017stochastic}      & 0.838        & 27.79       & 0.789       & 26.12      \\
          PredRNN  \cite{wang2017predrnn}    & 0.839        & 27.55       & 0.703     & 24.16    \\
          VarNet \cite{jin2018varnet}      & 0.843        & 28.48       & 0.739       & 25.37      \\
          SVAP-VAE  \cite{lee2018stochastic}   & 0.852        & 27.77       & 0.811       & 26.18     \\
          PredRNN++  \cite{wang2018predrnn++}  & 0.865        & 28.47       & 0.741     & 25.21    \\
          MSNET  \cite{lee2018mutual}      & 0.876        & 27.08       & --          & --          \\
          E3d-LSTM  \cite{wang2018eidetic}   & 0.879        & 29.31       & 0.810       & 27.24      \\
          STMFANet \cite{jin2020exploring}    & \underline{0.893}        & \underline{29.85}       & \underline{0.851}       & \underline{27.56}     \\ \midrule
          SimVP (ours) & \textbf{0.905}        & \textbf{33.72}       & \textbf{0.886}       & \textbf{32.93}           \\
          \bottomrule
      \end{tabular}}
  \label{table:KTH}
\end{table}

\vspace{-4mm}
\paragraph{Answer of Q3} SimVP extend well to the case of flexible predictive length. From Table.~\ref{table:KTH}, we know that SimVP achieve state-of-the-art performance. Notebly, the PSNR improves by 11.8\% and 19.5\% in both $(10 \rightarrow 20)$ and $(10 \rightarrow 40)$ settings. This phenomenon further indicates that the performance degradation of SimVP is less than others on long-term prediction tasks.


\vspace{-3mm}

\section{Ablation study}
While SimVP is quite simple, we believe there still exists unapprehended parts. We are eager to know:

\vspace{-2mm}
\begin{itemize}
  \setlength{\itemsep}{1pt}
  \item \textbf{Q1:} Which architectural design plays the key role in improving performance?
  \setlength{\itemsep}{1pt}
  \setlength{\parsep}{1pt}
  \setlength{\parskip}{1pt}
  
  \item \textbf{Q2:} How the Conv kernel influence the performance? 
  \setlength{\itemsep}{1pt}
  \setlength{\parsep}{1pt}
  \setlength{\parskip}{1pt}
  \item \textbf{Q3:} What roles do the Enc, Translator, and Dec play?
\end{itemize}
 
\vspace{-6mm}
\paragraph{Setting of Q1 and Q2} We perform the ablation study on Moving MNIST, TrafficBJ, and Human3.6. All models are trained up to 100 epochs, different from previous settings. For neural architecture design, we study whether using spatial UNet shortcut, temporal UNet shortcut, group convolution, and group normalization can bring performance gain. As to the convolutional kernel, we study how the kernel size and hidden dimension affect the model performance. We report the MSE metric for all datasets. 

\vspace{-3mm}
\paragraph{Answer of Q1} All of S-UNet, T-UNet, group convolution and group normalization can bring performance gain, and the order of significance is: $\text{group convolution} > \text{group normalization} \approx \text{S-UNet} \approx \text{T-UNet}$. Please see Table.~\ref{tab:ablation} (from model 1 to model 4) for experimental evidence.

\vspace{-3mm}
\paragraph{Answer of Q2} Larger kernel size and more model parameters lead to better performance. In Table.~\ref{tab:ablation}, from model 5 to model 8, with the increasing kernel size, we can see the significant performance gain. This improvement can be further enhanced by doubling the hidden dimension of model 8 to construct model 9. SimVP chooses multi-scale kernels, and the parameters of the Translator are 84\% of model 9. 


\vspace{-3mm}
\paragraph{Setting of Q3} We represent submodules trained with $n$ epochs as $\text{Enc}_{n}, \text{Translator}_{n}, \text{Dec}_{n}$, mix them and evaluate the results at $t=20$ (the last prediction) to reveal the role of each well-tuned module.

\begin{figure}[h]
  \centering
      \includegraphics[width=2.7in]{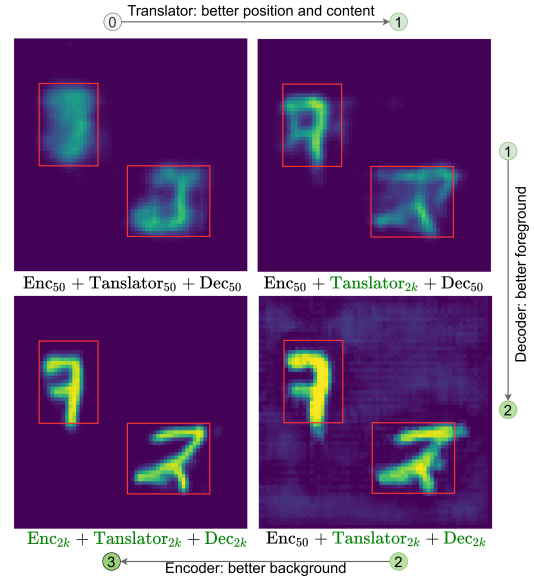}

    \caption{ The role of the Translator, Encoder and Decoder. }
    \label{fig:ablation}
\end{figure}

\vspace{-3mm}
\paragraph{Answer of Q3} As shown in Figure.~\ref{fig:ablation}, we conduct that the translator mainly focus on predict the position and content of the objects. The decoder is responsible for optimizing the shape of the foreground objects. The encoder can erase the background error by means of spatial UNet connection.

\vspace{-2mm}

\vspace{-2mm}
\section{Conclusion}
We propose SimVP, a simpler yet effective CNN model for video prediction. We show that SimVP can achieve state-of-the-art results without introducing any complex modules, strategies and tricks. Meanwhile, the reduced computing cost makes it easy to scale up to more scenarios. We believe simpler is better, and SimVP may serve as a strong baseline and provide inspiration for future research.

\vspace{-3mm}
\paragraph{Acknowledgemnet}

This work is supported in part by the Science and Technology Innovation 2030 - Major Project (No. 2021ZD0150100) and National Natural Science Foundation of China (No. U21A20427).

\clearpage
{\small
\bibliographystyle{ieee_fullname}
\bibliography{video}
}

\clearpage
\input{appendix.tex}

\end{document}

%% file: appendix.tex
\section{Appendix}
\subsection{Dataset}
\paragraph{Moving MNIST} Moving MNIST \cite{srivastava2015unsupervised} is a standard benchmark consisting of two digits independently moving within the 64 $\times$ 64 grid and bounced off the boundary. By assigning different initial locations and velocities to each digit, we can get an infinite number of sequences of length 20, thus enabling us to evaluate models without considering the data insufficiency issues. By default, models are trained to predict the future 10 frames conditioned on the previous 10 frames. Although the dynamics seem simple at first glance, accurately predicting consistent long-term future frames remains challenging for state-of-the-art models.

\paragraph{TrafficBJ} TrafficBJ contains the trajectory data in Beijing collected from taxicab GPS with two channels, i.e. inflow or outflow defined in \cite{zhang2017deep}. We choose data from the last four weeks as testing data, and all data before that as training data. Following the setting in \cite{wang2019memory}, we transform the data into $[0, 1]$ via max-min normalization. Since the original data is between -1 and 1, the reported MSE and MAE are 1/4 and 1/2 of the original data respectively after transformation, which is consistent with previous literature \cite{guen2020disentangling,wang2019memory}.

\paragraph{Human3.6} Human3.6 \cite{ionescu2013human3} is a complex human pose dataset with 3.6 million samples, recording different activities such as taking photos, talking on the phone, posing, greeting, eating, etc. Similar to \cite{srivastava2015unsupervised,wang2019memory,guen2020disentangling}, only videos with "walking" scenario are used. Following previous works, we generate 4 future frames given the previous 4 RGB frames.

\paragraph{KITTI\&Caltech Pedestrian} KITTI \cite{geiger2013vision} is one of the most popular datasets for mobile robotics and autonomous driving, as well as a benchmark for computer vision algorithms. It is composed by hours of traffic scenarios recorded with a variety of sensor modalities, including high-resolution RGB, gray-scale stereo cameras, and a 3D laser scanner. CalTech Pedestrian \cite{dollarCVPR09peds} is a driving dataset focused on detecting pedestrians. It is conformed of approximately 10 hours of $640 \times 480$ 30 FPS video taken from a vehicle driving through regular traffic in an urban environment, making a total of 250,000 annotated frames distributed in 137 approximately minute-long segments. We follow the same protocol of PredNet \cite{lotter2016deep} and CrevNet \cite{yu2019efficient} for preprocessing, training and evaluation. Models are trained on KITTI dataset  to predict the next frame after 10-frame warm-up and are evaluated on Caltech Pedestrian.

\paragraph{KTH} The KTH dataset \cite{schuldt2004recognizing} contains 25 individuals performing 6 types of actions, i.e., walking, jogging, running, boxing, hand waving and hand clapping. Following \cite{villegas2017decomposing,wang2018eidetic}, we use person 1-16 for training and 17-25 for testing. Models are trained to predict next 20 or 40 frames from the previous 10 observations.

\subsection{Translator: should we use RNN, Transformer or CNN?}

  \begin{figure}[h]
    \centering
        \includegraphics[width=3.2in]{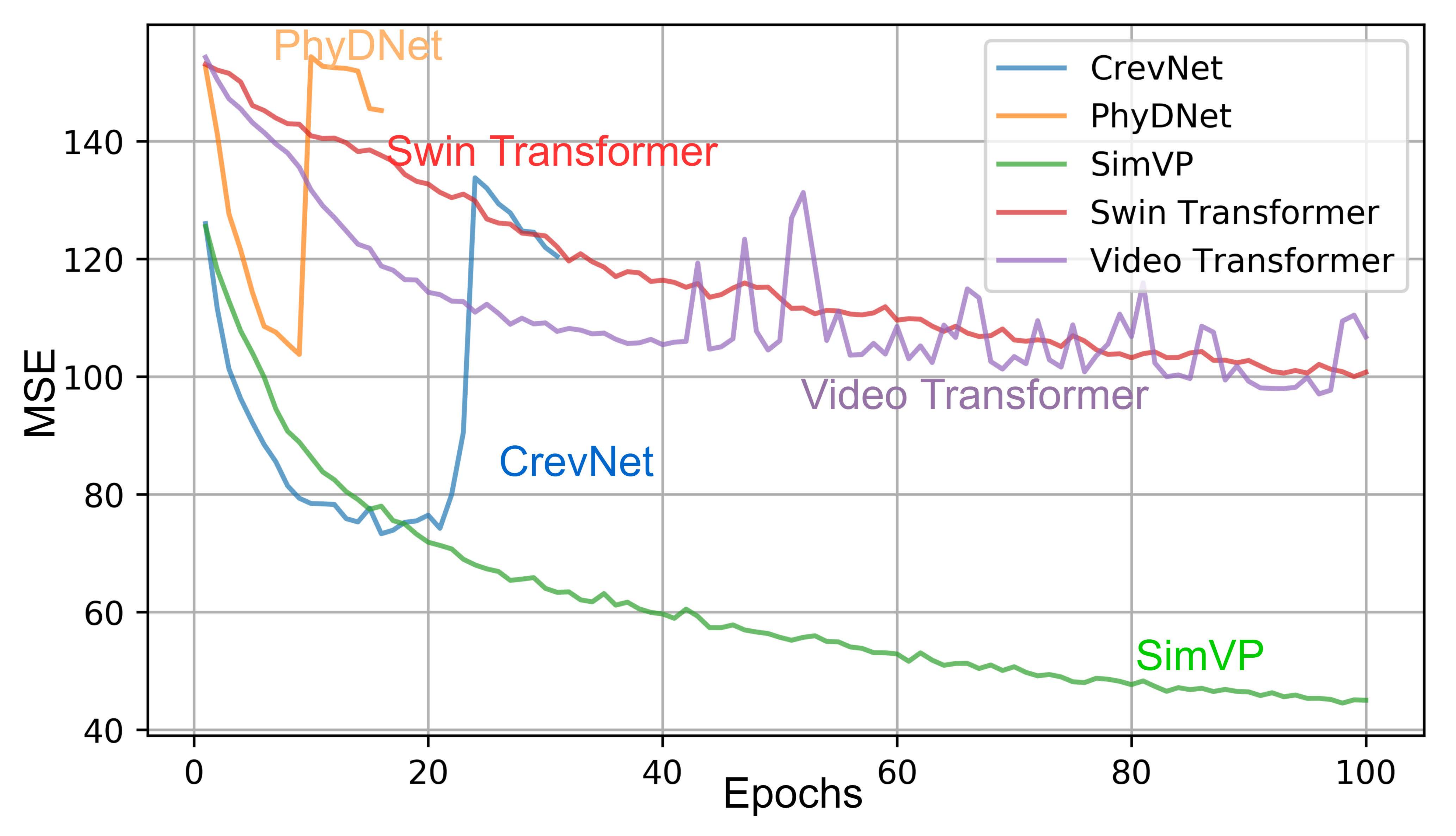}
      \caption{ Training dynamics of various translators on Moving MNIST using large learning rate 0.01.}
      \label{fig:cmp_translators_lr}
  \end{figure}

\subsection{Network structure}
Denote $N_{s}, C_{s}$ as the layer numbers and hidden dimensions of the spatial Encoder (or Decoder). Similarly, $N_{t}$ and $C_{t}$ are the layer numbers and hidden dimensions of the Translator's encoder (or decoder). We use NNI (Neural Network Intelligence) to search hyperparameters, and the search space is shown in Table.~\ref{tab:seach_space}.  The final hyperparameter settings on various dataset can be found in Table.~\ref{tab:hyper_param}.

\begin{table}[h]
  \centering
  \setlength{\tabcolsep}{4mm}{
  \begin{tabular}{cc}
  \toprule
          & Value              \\
  \midrule
  $C_{s}$  & 16,32,64    \\
  $C_{t}$ & 64,128,256,512 \\
  $N_{s}$      & 1,2,3,4     \\
  $N_{t}$   & 2,3,4,5 \\
  \bottomrule
  \end{tabular}}
  \caption{Search space}
  \label{tab:seach_space}
\end{table}

\begin{table}[h]
  \centering
  \begin{tabular}{cccccc}
    \toprule
          & Human & MMNIST & TrafficBJ & Caltech & KTH \\
  \midrule
  $C_{s}$  & 64    & 64     & 64        & 64      & 32  \\
  $C_{t}$ & 64    & 512    & 256       & 128     & 128 \\
  $N_{s}$      & 1     & 4      & 3         & 1       & 3   \\
  $N_{t}$      & 5     & 3      & 2         & 3       & 4  \\
  \bottomrule
  \end{tabular}
  \caption{Final hyperparameter setting.}
  \label{tab:hyper_param}
\end{table}




